\def\year{2020}\relax
\documentclass[letterpaper]{article} 
\usepackage{aaai20}  
\usepackage{times}  
\usepackage{helvet} 
\usepackage{courier}  
\usepackage[hyphens]{url}  
\usepackage{graphicx} 
\usepackage{graphicx}  
\usepackage{multirow}
\usepackage{amsmath,amssymb,amsfonts}
\usepackage{mathtools}
\usepackage{bm}      
\usepackage{dsfont}
\usepackage{xcolor}

\newcommand{\beginsupplement}{%
	\setcounter{table}{0}
	\renewcommand{\thetable}{S\arabic{table}}%
	\setcounter{figure}{0}
	\renewcommand{\thefigure}{S\arabic{figure}}%
}

\title{Seq2Emo for Multi-label Emotion Classification Based on Latent Variable Chains Transformation}
\author{Chenyang Huang, Amine Trabelsi, Xuebin Qin, Nawshad Farruque, Osmar R. Za\"{\i}ane\\
	Department of Computing Science, University of Alberta \\
	{\tt \{chuang8,atrabels,xuebin,nawshad,zaiane\}@ualberta.ca} \\}

\begin{document}

\maketitle

\begin{abstract}
Emotion detection in text is an important task in NLP and is essential in many applications. Most of the existing methods treat this task as a problem of single-label multi-class text classification.  To predict multiple emotions for one instance,  most of the existing works regard it as a general Multi-label Classification (MLC) problem, where they usually either apply a manually determined threshold on the last output layer of their neural network models or train multiple binary classifiers and make predictions in the fashion of one-vs-all.  
However, compared to labels in the general MLC datasets, the number of emotion categories are much fewer (less than 10). Additionally, emotions tend to have more correlations with each other. For example, the human usually does not express ``joy'' and ``anger'' at the same time, but it is very likely to have ``joy'' and ``love'' expressed together. Given this intuition, in this paper,  we propose a Latent Variable Chain (LVC) transformation and a tailored model -- Seq2Emo model that not only naturally predicts multiple emotion labels but also takes into consideration their correlations. We perform the experiments on the existing multi-label emotion datasets as well as on our newly collected datasets. The results show that our  model compares favorably with existing state-of-the-art methods. 
\end{abstract}

\section{Introduction}
Emotion mining from text \cite{SailunazDRA18,YadollahiSZ17} has attracted increasing attention in the recent research on Natural Language Processing (NLP) . However, most of the existing works regard this task as a problem of general \emph{multi-class} text classification. Multi-class 
text classification problem associate a single label $l$ from a set of single labels to any instance $X$. However, there are many other scenarios where an instance $X$ may have multiple labels. The detection of human emotions, for example, is one 
such scenario. Due to the complexity of human emotions, it is very likely that multiple emotions are expressed by a single text instance. 
These emotions may also be correlated. For example, emotions such as `hate' and  `disgust' 
may occur
more often together than in isolation. Typically, the number of possible expressed emotions is not large. 
Therefore, in this work, we regard emotion mining from the text as a special case of a Multi-label classification (MLC) problem
where the number of labels (emotions) is small, and where correlations may subsist between them. 

Common approaches to MLC problems usually involve various ways of \emph{problem transformation}, where an MLC problem could be transformed into multiple single-label text classification problems. Consequently, general single-label classifiers may be adopted directly or with modifications. Well known transformation techniques include Binary Relevance (BR) \cite{GodboleS04}, Classifier Chains (CC) \cite{ReadPHF11}, and Label Powerset (LP) \cite{tsoumakas2010random}. Given, $\mathcal{L} = \{l_1, l_2, \cdots, l_k\}$ 
a
set of labels, both transformations of BR and CC need to train $k$ binary classifiers where each of the classifiers is responsible for discriminating a single label $l_i$. Compared to BR, CC takes into account the correlations among the labels, whereas BR predicts each label independently. LP treats each possible combination of the labels as a separate label, therefore it may expand the number of labels to $2^k$ so that it is usually not feasible when $k$ is large.

In light of recent advances in Neural Network (NN) research which have shown great success in many NLP tasks  \cite{PetersNIGCLZ18,devlin2018bert}, we propose a new problem transformation which only uses the latent variables of a NN model as the ``chains'' to perform the task of MLC. As an analogy to CC, we refer to this transformation as Latent Variable Chains (LVC). 
Based on the proposed
LVC method, we also tailor 
a deep Neural
model -- \emph{Seq2Emo} which first captures both the semantic and the emotional features of an instance $X$, and then uses a bi-directional LVC to generate labels.
The model performs a sequence of predictions based on the chain of latent features which leads to
a multi-label emotion classification.
In addition, we 
collect a dataset that contains Balanced Multi-label Emotional Tweets (BMET), from scratch, to test both the baseline models and our proposed approach. 

The main contributions of this research are: (1) We propose an MLC problem transformation named \textbf{LVC} 
designed for NN models and more importantly, takes into account the correlations among target labels, 
which can be relevant for tasks like multi-label emotion classification.
(2) Moreover, we propose \textbf{Seq2Emo}, a novel NN model 
based on LVC,
that utilizes many recent research developments in deep learning and achieves encouraging results for classifying emotions in text.
(3) Furthermore, to validate the proposed methods, we assemble a new dataset, \textbf{BMET}, a large and balanced multi-label emotion dataset.
We make both the new dataset and the source code available to the public \footnote{The dataset and related code will be made publicly available after publication}.

In the remainder of the paper, we first present a brief synopsis of current  multi-label  classification  approaches in Section \ref{sec:related_work}. Then, in Section \ref{sec:overview}, we provide some preliminaries to better understand our model which is presented in Section \ref{sec:proposed}. The existing emotion text data benchmarks and how we gathered our own emotion text collection, are exhibited in Section \ref{sec:dataset}. We then present our experiments in Section \ref{sec:exp} and analyze the results in Section \ref{sec:results}. Perspectives and conclusions are highlighted in Section \ref{sec:conlusion}.
\section{Related Work}
\label{sec:related_work}


Multi-label classification (MLC)
assigns one or more labels to each sample 
in the dataset, as opposed to single-label classification which assigns a unique label to each sample.
In this section we present an overview of MLC methods in general. The MLC for emotion detection are mainly adaptations of the more general MLC approaches. 

One of the main approaches to MLC is the transformation based approach. It transforms an MLC task into some one-vs-all single labeling problems \cite{BOUTELL2004,ReadPHF11}.
Correlations or co-occurrences between labels is simply ignored in this case, as the problem is converted into isolated classification problems.
Several MLC models for emotion follow this approach, transforming the problem into a 
binary classification problem \cite{Baziotis2018SemMulti}.
However, these models can be computationally expensive when using large amount of labels or datasets.

Another set of methods 
applies the threshold dependent approach.
The methods usually set a threshold on the output probabilities in order to determine the predicted classes
\cite{chen2017ensemble,kurata2016}.
In our work, we
consider multi-label emotion classification as a fixed length label-sequence generation problem.
Instead of a threshold dependent model, 
the labels of emotion are generated sequentially, and are dependent of each other.



Recently deep learning methods for emotion classification have exhibited success.
Using deep learning methods 
allows to avoid the labor-intensive task of feature engineering that is usually necessary with other classification paradigms \cite{yan2016}. 
Deep learning methods also
propose an end-to-end framework for classification, but remain dependent on a threshold function that needs to be learned or implemented \cite{Yu2018emnlp}.
Finding 
a good threshold function is a challenging problem in itself \cite{fan2007study}. 
Other methods like \citeauthor{he2018joint} (\citeyear{he2018joint})'s work , although not using any threshold transformation based approach, they incorporate some prior knowledge on the different emotion relations for a better classification.
In this work, we do not use any external or prior knowledge, or a threshold mechanism. 







\section{Overview}
\label{sec:overview}
In order to compare our proposed LVC transformation with both BR and CC methods, in this section, we systematically introduce both BR and CC transformation methods. We also explain how NN models can be transformed using the two methods.

\subsection{The MLC Task}
In text classification tasks, an instance $X$ is usually in the form of $X = [x_1, x_2, x_3, \cdots, x_n]$  where $x_i$ is a word or token and $n$ is the length of the sequence. In addition, each $X$ is assigned with a target $Y$, where $Y  \subseteq \mathcal{L}$ represents the corresponding labels of $X$. The set relation ``$\subseteq$''  indicates that each $Y$  may contain multiple elements in $\mathcal{L}$ or none of them (i.e. $Y = \emptyset$).

Hence, an MLC model 
is supposed to learn the conditional distribution of $\mathcal{P}(Y|X)$, where $Y$ is a set and the number of elements $|Y|$ is not always equal to one. 


\subsection{BR Transformation}
\label{sec:br_trans}
\emph{Binary relevance} transformation is a simple but very adaptive method which allows 
the incorporation of any single-label classifier
to
the task of MLC. 

To begin with, the target $Y$ is represented as a binary vector $Y^b = (y_1, y_2, \cdots, y_k)$, where $y_i \triangleq \mathds{1}(l_i \in  Y)$ \footnote{The symbol ``$\triangleq$'' reads as ``is defined as'', and it is not the same as ``$=$'' which indicates  ``equal to''. $\mathds{1}$ is the indicator function, its value will be equal to 1 if the inner condition is satisfied, otherwise the value will be 0.}.
In general, when the size of the label set $\mathcal{L}$ is $k$, $k$ individual models will be required for this type of transformation. Denote the classifiers for BR transformation are $C_j^{B}$ ($j \in [1 \cdots k]$).
The classifier $C_j^{B}$ is only responsible for the generation of $y_i$. In 
other
words, $C_j^{B}$ is learning the probability distribution of $\mathcal{P}(y_i|X)$,
and
$Y^b$ is generated by the predictions of all $k$ classifiers.

Traditional classifiers, such as SVM, Na\"{i}ve Bayes, etc. can be used for MLC tasks by BR transformations. The deep learning models, although can still be adopted directly, do not necessarily need $k$ totally individual models. When applying NN models for single-label text classification, instance $X$ is firstly represented by some types of NN encoders, such as CNN \cite{kim2014convolutional}, RNN \cite{HochreiterS97}, Transformer \cite{vaswani2017attention}, etc., and then apply a Fully Connect (FC) layer to project the representation to the space of labels using \emph{SoftMax} normalizer. The FC layer itself can be regarded as a classifier which takes as input the vector representation of $X$ which is generated by any encoders. Therefore, it is more efficient for the $k$ classifiers to share a same encoder and only have different FC layers. 

\begin{figure}[ht]
	\centering
	\includegraphics[width=0.28\textwidth]{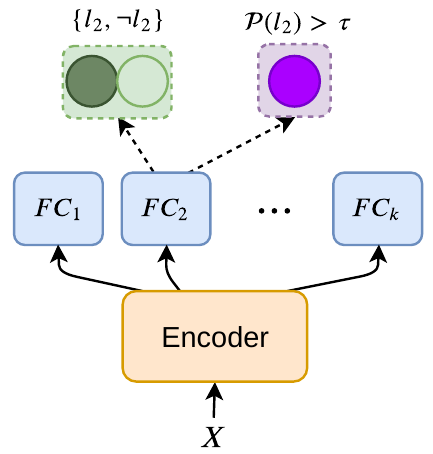}
	\caption{Two variants of Binary Relevance (BR) transformed NN models. The top left (green) block indicates the binary classifier using two cells, and the top right (purple) block indicating a classifier using one cell and threshold $\tau$}
	\label{fig:br_nn}
\end{figure}

Given, FC layers denoted as $FC_j$ ($j \in [1, \cdots, k]$), in the case of BR transformation, each $FC_j$ is responsible for a binary classification. As shown in Figure \ref{fig:br_nn}, there are two major variants of binary FC classifiers that can be used. The first type of the classifier has two cells as the other end of the FC layer. The outputs of the two cells are regularized by \emph{SoftMax}. Denote the two cells $b_j^0$ and $b_j^1$, then $y_i \triangleq \mathds{1}(b_j^0 < b_j^1)$. Another method involves an additional hyper-parameter $\tau$ and there is only one cell on the other end of the $FC_j$ layer. The output is usually regularized by \emph{Sigmoid} activation function so that the output will be in the range of $(0, 1)$. Correspondingly, denote the output as $b_j^r$ and  the binary classification is achieve through $y_i \triangleq \mathds{1}(b_j^r > \tau)$.

In this models, it is assumed that the labels are independent. However, this assumption usually does not hold, especially when the labels 
correspond to 
emotions \cite{shahraki2017lexical}. 

\subsection{CC Transformation}

In order to take into consideration the correlations of the labels in $\mathcal{L}$, \citeauthor{ReadPHF11} (\citeyear{ReadPHF11}) proposed another transformation method called \emph{classifier chains}. 

Similar to BR, CC transformation also requires $k$ individual classifiers. Given, the classifiers, $C_j^C$, where $j \in [1 \cdots k]$. The original CC transformation conducts $k$ continues binary classifications where each classification is based on the output of the previous one.

Using the binary representation $Y^b$, the transformation can be represented as the following recursive procedure:
\begin{equation}
\label{eq:cc}
y_{j} = C^C_j(X, y_{j-1}),
\end{equation}
Where $y_0 = \emptyset$ (i.e. $C^C_1$ only take as input $X$).

When CC is firstly proposed, DL methods was not as popular as the time of conducting this research. The use of CC transformation was restricted by the traditional models which are not as flexible on inputs and output as that on NN models in general. In the following 
paragraphs,
we explain how NN models 
can use
CC transformation methods.

In fact, Seq2Seq model \cite{sutskever2014sequence} which is widely used for neural machine translation, document summarization and end-to-end dialogue generation, is adopting the very similar idea of CC transformation. It usually contains two major components: \emph{encoder} and \emph{decoder}.

The encoder compresses the information of the sequence $X$ into dense vector/vectors representations:
\begin{equation}
\label{eq:enc}
\bm{v} = Encoder(X)
\end{equation} 

Given $\bm{v}$, the decoder normally predicts the target $Y^b$ sequentially using the following formula: 
\begin{equation}
\label{eq:dec}
y_{j} = Decoder(\bm{v}, y_{j-1}).
\end{equation}
where $y_0$ is usually a special token \textless s\textgreater ~to indicate the start of the decoding.
By comparing Eq. \ref{eq:cc} and \ref{eq:dec}, one may find they are very similar to each other. 
The 
only
major difference is that the decoder in Seq2Seq is a single model whereas the original CC method requires $k$ individual models.

To the best of 
our
knowledge,
there has not been any existing research that directly uses Seq2Seq model upon the binary representation $Y^b$.
\citeauthor{YangCOLING2018} (\citeyear{YangCOLING2018}), however, applied Seq2Seq model on a different representation of the target label set $Y$. Their system is named 
\emph{SGM} (Sequence Generation Model for Multi-label Classification). In their approach, $Y$ is transformed into an ordered sequence $Y^o$. Once an arbitrary order of the full label set $\mathcal{L}$ is determined (denoted
as $\mathcal{L}^o$). All the elements in $Y^o$ will occur from left to right using the same order as that in $\mathcal{L}^o$. 

The CC transformation has a critical problem.
In the inference phase, the target sequence $Y$ is unknown. 
Thus,
applying
Eq. \ref{eq:cc}
or
\ref{eq:dec} 
is not possible as
the \emph{true} $y_{j-1}$ 
is unknown
which prevents
the generation of
the next label.
However, 
it is possible to use
the estimated  $y_{j-1}$ 
instead, in order 
to continue the iteration. Therefore, if the model use the true $y_{j-1}$ rather than estimated label in the training phase (which is also known as \emph{teach forcing}),  there will be an inconsistency between the training and the inference phases. \citeauthor{bengio2015scheduled} (\citeyear{bengio2015scheduled})  refer to this issue as \emph{exposure bias}. We propose a new transformation method that circumvents this problem.

\begin{figure*}[ht]
	\centering
	\includegraphics[width=0.98\textwidth]{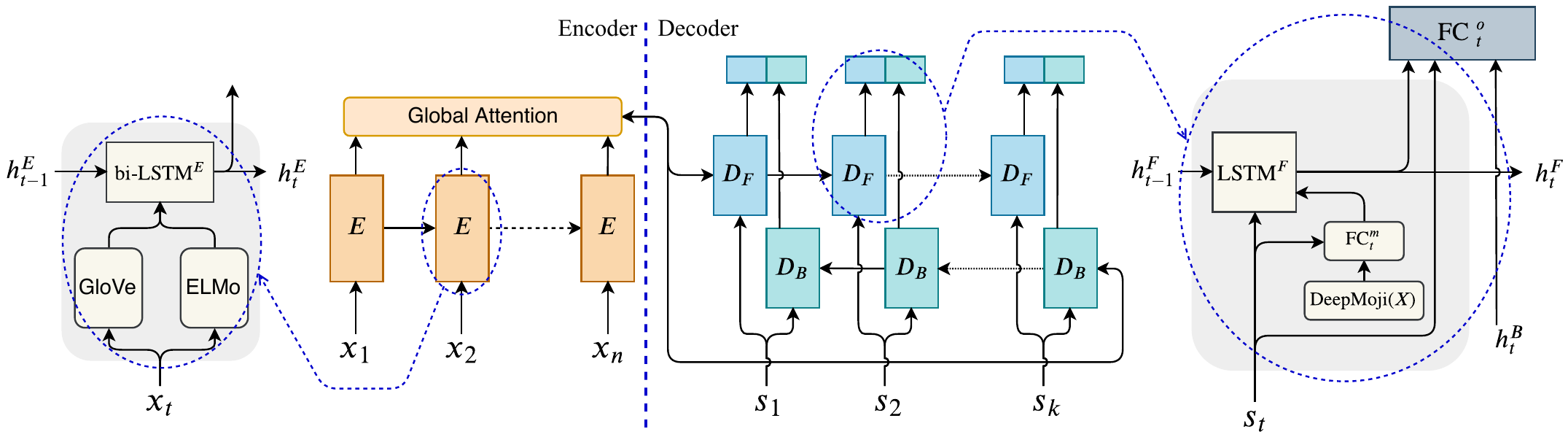}
	\caption{Overview of the Seq2Emo model}
	\label{fig:seq2emo}
\end{figure*}

\section{Proposed System}
\label{sec:proposed}
In the proposed model, we also have an encoder and a decoder. However compared to SGM, we propose a different problem transformation of MLC tasks.  
For SGM, the length of the target sequence $Y$ is dependent on the number of elements in the set of positive emotions $y$ and, therefore, it is directly dependent on the previous predicted label. Hence, the issue of exposure bias is inevitable by 
SGM's
problem setup. In this section, we first propose a different transformation scheme which does not explicitly use $y_{j-1}$ to connect the ``chain'', 
and thus avoids the problem of exposure bias.

\subsection{LVC Transformation}
\label{sec:lvc}
We propose a transformation method called Latent Variable Chains (LVC). LVC make use of the special property of Recurrent Neural Network (RNN) models which have  intermediate hidden states at each step of decoding. 

LVC uses the same label representation $Y^b$ as that in BR transformation, in addition, it requires a sequence of signals $S = (s_1, \cdots, s_k)$. The signal has the same length 
as
the size of total label set $\mathcal{L}$ and each signal is an auxiliary input for generating a binary label in $Y^b$.

Formally, LVC first use the same encoder as that in Eq. \ref{eq:enc}, and its decoder is modified as the following:
\begin{align}
h_{j} & = Decoder(\bm{v}, h_{j-1}, s_{j})  \label{eq:lvc_dec_hidden} \\ 
y_{j} & = Proj (h_j) \label{eq:lvc_dec_proj}  
\end{align}
In Eq. \ref{eq:lvc_dec_hidden}, $h_j$ is the hidden state from an RNN decoder, which is further used to generate $y_j$. Compared to Eq. \ref{eq:dec}, $y_{j-1}$ is not required to generate $y_j$. Instead, we use the signal $s_j$ as the auxiliary input to assist the decoding of $y_j$.
Thus,
there will be no exposure bias problem during
training and inference phases.
Additionally, the hidden state $h_j$ still retain the information of the previous decoding step 
(via input $h_{j-1}$),
so that the correlations 
among
the labels can be calculated in the process of generating the binary label $Y^b$. $Proj$ is a model that project the hidden states into a binary space (two cells) 
in order to be 
able to estimate the likelihood of label $y_j$
using \emph{SoftMax}. 
The common choice of a $Proj$ layer can be one or multiple \emph{dense or fully connected} layers. 

However, 
one limit of this setting is that
the correlations of labels are considered in only one direction, which might be sub-optimal. For example, say $y_3$ stands for emotion ``angry'' and $y_4$ is corresponding to ``sadness''. Eq. \ref{eq:lvc_dec_hidden} suggests that while recognizing the emotion ``sadness'', the latent variable which is used to detect ``angry'' is considered, but this information flow does not exist the other way around. In this regard, in order to fully consider the relations among the emotions, a model 
of
chained classifiers needs to
diversify the order of emotion labels.
\citeauthor{ReadPHF11} (\citeyear{ReadPHF11}) have used ensembles of different CC classifiers with different orders 
to tackle this issue.
However, they were 
limited by the lack of the ability of
traditional classifiers
to learn
complex relations, 
as well as their
lack of flexibility.
RNN models such LSTM \cite{schuster1997bidirectional} and GRU \cite{gru} are able to 
effectively
capture the relatively long distance relations of a sequence in one direction. Stacking only two RNN models of two reversed orders is intuitively a good solution. This idea is also known as  bidirectional RNNs \cite{schuster1997bidirectional} which has already been widely used in many applications \cite{huang2015bidirectional}. 
Following the same idea, we further 
extend LVC transformation into a bidirectional fashion as follows:
\begin{align}
	\overrightarrow{h}_{j}	& = Decoder^F(\bm{v}, \overrightarrow{h}_{j-1}, s_{j}) \label{eq:bi_lvc_f} \\ 
	\overleftarrow{h}_{j} 	& = Decoder^B(\bm{v}, \overleftarrow{h}_{j-1}, s_{k-j + 1})	 \label{eq:bi_lvc_b}  \\
	y_j 				  	& = Proj(\overrightarrow{h}_{j}, \overleftarrow{h}_{k-j + 1}) \label{eq:bi_lvc_p}
\end{align}
where $Decoder^F$ and $Decoder^B$ stands for forward decoder and backward decoder respectively.

\subsection{Seq2Emo Model}




In Section \ref{sec:lvc}, we have 
presented
the \emph{Latent Variable Chain} transformation for MLC task. In this section, we introduce the Seq2Emo model which is not only tailored to adapt the proposed transformation, but also dedicated for the task of emotion classification.

Figure \ref{fig:seq2emo} shows the high level overview of Seq2Emo, where $\text{E}$ stands for encoder, $\text{D}_F$
and $\text{D}_B$ stand for a forward decoder and a backward decoder respectively. Seq2Emo also contain many other modules: ``GloVe'', ``ELMo'', ``Global attention'', and ``DeepMoji'', which we will explain their use in the later part of this section.
The design of Seq2Emo is based on the existing Sequence-to-Sequence (Seq2Seq) model \cite{sutskever2014sequence}
and its \emph{encoder-decoder} structure. 
However, the
two 
models are
different 
as they are
adapting two different transformations: 
Seq2Emo is adapting LVC 
while Seq2Seq
is 
adapting
CC.


\subsubsection{Encoder}
We show the structure of the encoder in the left side of Figure \ref{fig:seq2emo}, where we use a multi-layer bi-LSTM to encode the input sequence $X = [x_1, x_2, x_3, \cdots, x_n]$. Inspired by the work of \cite{sanh2018hierarchical},
we 
use the combination of GloVe \cite{pennington2014glove} and ELMo \cite{PetersNIGCLZ18} to best capture the local semantic meaning and the contextual representation of each word $x_t$. 

\paragraph{GloVe and ELMo}
GloVe and ELMo are both pre-trained models which have been trained on large text corpora using unsupervised learning algorithms. GloVe does not differentiate the meaning of a word given its context, instead, it uses the probability distribution of a word's context to represent its semantic meaning. 
Similar approaches exist in the literature \cite{mikolov2013efficient,bengio2003neural,joulin2016fasttext}. 
However, in 
our
proposed framework, we found GloVe to work slightly better than the others. ELMo is a 
recent RNN based contextual word vector model. Unlike GloVe, it assigns a dense vector representation to each word dynamically based on its context. Besides, ELMo is trained at the character level which can be used to capture the semantic meaning of out-of-vocabulary words. Therefore, we combine the GloVe and ELMo as the feature representation of the words.

Given, $Glv$ as a pre-trained GloVe model, the word vector for $x_t$ generated by GloVe can be thus represented as $Glv(x_t)$. $Glv$ is a matrix of dimension $\mathbb{R}^{|V| \times {D_G}}$, where $D_G$ is a hyper-parameter of a GloVe model and each word in GloVe will be represented in a vector space of $\mathbb{R}^{D_G}$. $V$ is the set of words that Seq2Emo is able to recognize, $|V|$ is the number of words. 

To generate the vector representation of each word $x_t$, a pre-trained Elmo model, $Elm$, first takes as input $X$, and generate a matrix of dimension $\mathbb{R}^{|n| \times {D_E}}$, where $n$ is the length of the sequence $X$ and ${D_E}$ is a hyper-parameter of the pre-trained ELMo model. Given $X$, we denote the generated matrix by the ELMo model as $Elm^X$. Further more, we use a subscript $t$ to the to denote the $t^{th}$ row of $Elm^X$. Therefore, the word vector of $x_t$ given by $Elm$ can be written as $Elm_t^X$ and each vector will be in the space of $\mathbb{R}^{D_E}$.

\paragraph{LSTM encoder} After having both vector representations of a word $x_t$, denote $[Glv(X_t);Elm_t^X]$ as the concatenation of the two vectors. By using the concatenation we have two pre-trained word representation models combined, and each word $x_t$ is represented in a vector space of $\mathbb{R}^{D_G +D_E}$. We then use a multi-layer bi-directional LSTM  \cite{deeprnn} to encode the concatenated vectors. For simplicity, we only use $\text{bi-LSTM}^{E}$ to denote the model, and its iterations can be shown as:
\begin{equation}
\label{eq:1}
	h_t^{E}, c_t^{E} = \text{bi-LSTM}^{E}([Glv(X_t);Elm_t^X], [h_{t-1}^{E}; c_{t-1}^{E}]),
\end{equation}

where $h_0^{E} = c_0^{E} = \bm{0}$ and $h_t^{E}$ and  $c_t^{E}$ represent the \emph{hidden} state and \emph{cell} state of the LSTM model at time step $t$. We further use $\bar{h}_t$ and $\bar{c}_{t}$ to denote the states of the top layer of the deep LSTM model.

\paragraph{Global attention}
In this research, we use the \emph{global attention}  \cite{luong2015effective} to better capture the long-distance dependency between the encoder and decoder in the Seq2Seq framework.
More specifically, we use the \emph{general} alignment score function and \emph{input-feeding} update scheme. Given $\bm{\bar{h}}^{E} = [\bar{h}_1^{E}, \bar{h}_2^{E}, \cdots, \bar{h}_n^{E}] $ as all the hidden states of the top LSTM layer from the encoder. 
 We choose a single-layer single-directional LSTM for each of the \emph{forward} and \emph{backward} decoding direction. Denote $h^D_t$ as the hidden state of time step $t$ of a decoder $D$. It will be updated by the attention mechanism through  $h^D_t ~ \rightarrow ~ \bm{\alpha}_t ~ \rightarrow ~ CTX_t ~ \rightarrow ~ \tilde{h}^D_t $. $\bm{\alpha}_t$ is usually referred as the vector of \emph{attention scores}. 
$CTX_t $ is the \textit{context vector} which is dynamically calculated at each decoding step $t$, it corresponds to a vector which is the weighted sum of encoder output $\bm{\bar{h}}^{E}$ and the attention scores $\bm{\alpha_t}$. The updating of decoder hidden state $h^D_t$ using global attention is shown in the following iterations:

\begin{align}
\tilde{{h}}^D_{t} & =\tanh \left({W}_{{c}}\left[{CTX}_{t} ; {h}^D_{t}\right]\right) \\
CTX_t & = \frac{  \sum \bm{\alpha_t} \bm{\bar{h}}^E   }{\sum \bm{\alpha_t}} \\
 \bm{\alpha_t}(i) & = \frac{\exp \left(\operatorname{score}\left(h_t^{D}, \bar{h}_i^{E} \right)\right)}{\sum_{j=1}^n \exp \left(\operatorname{score}\left(h_t^{D}, \bar{h}_j^{E} \right)\right)} \\
 \text{score} \left({h^D_t}, \bar{h}_i^E \right) & = {{h^D_t}^{\top} {W}_{{a}}  \bar{h}_i^E}  
\end{align}

where $\bm{\alpha_t} = (\alpha_1, \alpha_2, \cdots, \alpha_n)$ and  $ \bm{\alpha_t} (i) \triangleq \alpha_i$.

\subsubsection{Decoder}
The design of the \emph{decoder} is the core of the proposed Seq2Emo model. Its graphical representation is shown 
on 
the right side of the Figure \ref{fig:seq2emo}.
We
can 
observe
two different FC modules and a module named  ``DeepMoji'' along side the decoder LSTM. We can also
notice
two decoders of 
reversed directions: $D^F$ and $D^B$.
We 
explain the forward decoding of Seq2Emo in details in this section. It can be easily adopted to backward decoding by changing the subscript 
similarly to
what 
has been done
in Eq. \ref{eq:bi_lvc_b} compared to Eq. \ref{eq:bi_lvc_f}.

DeepMoji \cite{felbo2017} is an LSTM based model which is pre-trained on over 1 billion tweets by the task of predicting contained emojis. Given a sequence $X$, DeepMoji can give the emotional representation by extracting the output from the last layer. 
Given that the target domain of our study is
emotion classification, we plug DeepMoji encodings into the proposed Seq2Emo 
as it brings an emotional semantic dimension to the model. We denote this
process 
as $\text{Moji}_X = \text{DeepMoji}(X)$. Furthermore, we add a fully connected layer $FC^m_t$ where $t$ is depending on the signal $s_t$. This FC layer will project the static representation $\text{Moji}_X$ to a smaller but learnable vector while conditioning on the emotion that is about to be predicted. 

For a single direction of decoding, we use the following equations to update the hidden states:
\begin{align}
\label{eq:3}
h_t^{D}, c_t^{D} & = \text{LSTM}^{D}\Big([s_t, FC^m_t(\text{Moji}_X)], \big[\tilde{h}^{D}_{t-1}; c_{t-1}^{D}\big]\Big)  \\
h
^{D}_{0} & = \bar{h}^{En}_n , c_0^{D} = \bar{c}^{E}_n.
\end{align}

Denote $\overrightarrow{h}_t^{D}$ and $\overleftarrow{h}_t^{D}$ as the hidden states of forward decoder and backward respectively. We further apply different FC layers for each of the binary label classification in $Y^b$ similar to that in Section \ref{sec:br_trans}. This procedure can be represented as follows:
\begin{equation}
    y_t = FC_t^o([\overrightarrow{h}_t^{D}; \overleftarrow{h}_t^{D}])
\end{equation}

In the process of decoding, we feed a sequence of signals $S = (s_1,, \dots, s_{k})$ to the decoder to force a generation of $k$ labels. The sequence $S$ is dependent on size of the label set $\mathcal{L}$ which is a constant value for a given MLC task. 
%
Feeding the signals also means that in the training, each signal is telling the decoder to learn a specific label given the input sequence $X$. 
To feed a signal into LSTM directly, we modify the $\text{LSTM}^D$ in a similar way to 
that of 
\citeauthor{li2016persona}'s (\citeyear{li2016persona}).
Signal $s_t$ are used in both $FC^o_t$ and $FC^m_t$ so that each FC layer is only responsible for one emotion. During the experiments, we found that disentangling FC layers based on different labels is able to achieve better results in Seq2Emo model comparing to sharing a single FC layer.






\section{Data Collection and Pre-processing}
\label{sec:dataset}
Due the to lack of study in the area of multi-label emotion classification, 
the 
publicly accessible datasets
for this specific task 
remain rare. 
We first introduce two existing datasets that contain multi-labeled emotions and then explain the procedure of collecting a new large and balanced dataset with respect to each emotion category.

\begin{table}[ht]
	\caption{Number of the emotions, instances and the proportions of the multi-labled instances of the three datasets}
	\label{tab:dataset_stat}
	\centering
	\resizebox{0.45\textwidth}{!}{
		\begin{tabular}{l|ccc} \hline \hline
			Dataset   & \# of emotions & \# of instances & \% multi-label \\ \hline
			SemEval18 &   11      &   10690      & 86.1 \% \\
			CBET      &   9      &   81162          & 5.6  \% \\
			BMET      &   6      &   96323         & 24.2 \%  \\ \hline \hline
	\end{tabular}}
\end{table}








\subsection{SemEval18}
In the shared \emph{SemEval-2018 Task 1: Affect in Tweets} \cite{MohammadBSK18},
the provided dataset
is labeled by human annotators. 
It has been widely used. 
The task itself is composed of many sub-tasks, among which, \emph{E-c} is a task of emotion classification. It contains 11 emotion categories: ``anger'', ``anticipation'', ``disgust'', ``fear'', ``joy'', ``love'', ``optimism'', ``pessimism'', ``sadness'', ``surprise'', and ``trust''. 
This dataset consists 10,690 Tweets which are mostly multi-labeled (see Table \ref{tab:dataset_stat}). For simiplicity, we refer to this dataset as \emph{SemEval18} in the following context.

\subsection{CBET}
Unsing hashtags as self-annotated labels, \cite{shahraki2017lexical} created a tweet-based dataset and named it as Cleaned Balanced Emotional Tweets (CBET). It contains 9 emotions which are chosen from the union of four highly regarded psychological models of emotions.
The CBET dataset is much larger than SemEval18, however, very few instances are multi-labelled, most of the instances (94.4\%) are singly labelled.

\subsection{Balanced Multi-label Tweets (BMET)}

The SemEval18 dataset encompasses 11 emotions and has only a little more than ten thousand instances.
On the other hand, CBET is much larger but it is mostly single labeled. 
We collect a new dataset and name it Balanced Multi-label Tweets (BMET). It is larger than CBET dataset and contains much more multi-labeled instances. In addition, BMET is balanced with respect to each emotion category. The statistics are shown in Table~\ref{tab:dataset_stat}. BMET data is collected from scratch mostly following the procedure described in \cite{abdul2017emonet}. The details of the data collection, post-processing and samples can be found in the supplementary materials.

\begin{table*}[ht!]
\caption{Results}
\label{tab:results}
\centering
\begin{tabular}{l|ccc|ccc|ccc}  \hline \hline
\multirow{2}{*}{Models}&          & Semeval18 &          &          & CBET   &          &          & BMET   &          \\ \cline{2-10}
            & Jaccard & HL        & Micro F. & Jaccard & HL     & Micro F. & Jaccard & HL     & Micro F. \\ \hline
SGM         & 0.4514   & 0.1668    & 0.5511   & 0.5184   & 0.1090 & 0.5270   & 0.5689   & 0.1610 & 0.6052   \\
Binary-SL   & 0.5737   & 0.1230    & 0.6914   & 0.5573   & 0.0750 & 0.6368   & 0.5776   & 0.1266 & 0.6661   \\
Binary-SLD  & 0.5752   & 0.1231    & 0.6909   & \textbf{0.5823}  & \textbf{0.0739} & \textbf{0.6508}   & 0.5685   & \textbf{0.1258} & 0.6619   \\
Seq2Emo-L  & 0.5884   & 0.1217    & 0.7063   & 0.5750   & 0.0771 & 0.6433   & 0.5715   & 0.1301 & 0.6607   \\
Seq2Emo-LD & \textbf{0.5919}   & \textbf{0.1211}    & \textbf{0.7089}   & 0.5779   & 0.0766 & 0.6440   & \textbf{0.5856}   & 0.1276 & \textbf{0.6719}    \\ \hline \hline
\end{tabular}
\end{table*}

\section{Experiments}
\label{sec:exp}
\subsection{Metrics}
Following the work of \citeauthor{YangCOLING2018} (\citeyear{YangCOLING2018}) and \citeauthor{MohammadBSK18} (\citeyear{MohammadBSK18}), we  choose \emph{Jaccard Index} \cite{rogers1960computer}, \emph{Hamming Loss}  \cite{SchapireS99}, and \emph{Micro-averaged F1 score} \cite{Manning2008} as the automatic metrics. 

\paragraph{Jaccard index}
Jaccard index is also refferred as multi-label accuracy \cite{MohammadBSK18}. Denote the test set as $\{X^{te}_i, Y^{te}_i\}^N$,  where  each $Y^{te}_i$ is a set of emotion labels  and $N$ is the size of the test set. Let $\hat{Y}^{te}_i$ be the estimated labels by a model. Jaccard index can be defined as follows:

\begin{equation}
J =\frac{1}{N} \sum_{i=1}^N \frac{|Y^{te}_i \cap \hat{Y}^{te}_i|}{|Y^{te}_i \cup \hat{Y}^{te}_i|}
\end{equation}

It has to mention that in Semeval18 dataset, there are instances with $Y^{te}_i = \emptyset$, therefore it is possible that the value of $|Y^{te}_i \cup \hat{Y}^{te}_i| = 0$. In this situation, we regard the value of the corresponding term as 1, because  the estimated set $\hat{Y}^{te}_i$ corresponds to the true set $\emptyset$ in this case.

\paragraph{Hamming loss}
Hamming Loss (HL), is used to find out the number of wrongly predicted labels out of all labels. Hence, the less the score, the better.

\paragraph{Micro F1}
We refer Micro-averaged F1 score as \emph{Micro F} for simplicity. It takes into consideration of true positives, false negatives, and false positives, which is a widely used metrics for multi-label classification problems.





\subsection{Baseline models}

In this research, we propose a 
new
problem transformation 
approach,
LVC, 
for multi-label emotion classification.
We compare 
our proposed
LVC-based
model, Seq2Emo, against models that are based on the BR transformation and CC transformation (see Section \ref{sec:overview}). For BR transformation, we use the $SL$ and $SLD$ models from \cite{huang2019ana} as encoders, and adapt the models to solve MLC problems by adding multiple binary FC classifiers to the end (as shown in Figure \ref{fig:br_nn}). We name the models as Binary-SL and Binary-SLD respectively.
SL and SLD utilize many recent advances in text classification and emotion mining. However, in the research of \cite{huang2019ana}, SLD is part of the proposed hierarchical framework and its individual performance is not given. In this work, SL without self-attention is used as the encoder of the proposed Seq2Emo model, the DeepMoji Module is used by the decoder. In order to justify the the performance of SL and SLD models, we test them with three single label classification datasets, the results and analysis are shown in
Supplementary Material.
For CC transformation, we directly use the public implementation by the author of the SGM model \cite{YangCOLING2018} \footnote{\url{https://github.com/lancopku/SGM}}. To show the impact of DeepMoji module, we make two variants of the Seq2Emo model: Seq2Emo-L and Seq2Emo-LD. Seq2Emo-LD is the full sized model as shown in the Figure \ref{fig:seq2emo} and Seq2Emo-L simply removes the  DeepMoji module and its connection to the decoder LSTM.


\subsection{Hyper-parameters}
\label{sec:setup}
We use PyTorch 1.0
as the deep learning framework. For the DeepMoji model, we use the implementation offered by Hugging Face team\footnote{\url{https://github.com/huggingface/torchMoji}}. For the dimensions of the Bi-LSTMs encoders in Binary-SL, Binary-SLD, Seq2Emo-L, and Seq2Emo-LD, we set the dimension in each LSTM direction as 1,200. The number of layers of Bi-LSTM modules are set to 2. We use Adam optimizer with 5e-4 as the learning rate for Binary-SL, Binary-SLD and the encoder part of the proposed Seq2Emo-L and Seq2Emo-LD model. As for the decoder of the Seq2Emo-L and Seq2Emo-LD, we decrease the learning rate to 1e-4. We apply a Dropout rate \cite{srivastava2014dropout} of 0.2 to all the models. 

%
%
%


\section{Results and Analysis}
\label{sec:results} 

Table \ref{tab:dataset_stat} highlights the fact that SemEval18 has the highest multi-label percentage followed by BMET.  CBET has the lowest percentage, meaning that BMET, the dataset we collected, has more tweet posts with multiple labels. 
Table \ref{tab:results} shows the results of our methods and the contenders. 
As we can see, our model Seq2Emo-LD achieves better performance than 
contenders
on the datasets (Semeval18 and BMET), which have the highest percentages of multi-label samples.  
On CBET dataset, which is mostly single-labeled, Binary-SLD is able to achieve better performance. We show that  Binary-SLD is a very strong baselline model on single-label emotion classification task in the supplementary materials.
In addition, models with DeepMoji (Binary-SLD and Seq2Emo-LD) achieves obvious performance improvements on datasets Semeval18 and CBET against those models without using DeepMoji (Binary-SL and Seq2Emo-L).

The results generated by Seq2Emo-LD on Semeval18 dataset outperform the top solutions of the shared task \cite{park2018plusemo2vec}.  To the best of our knowledge, Seq2Emo model achieves the best Jaccard score on the SemEval18 dataset \cite{MohammadBSK18}.  



\section{Conclusion}
\label{sec:conlusion}
In this research, we aim at tackling the task of multi-label emotion classification. 
We 
propose LVC  transformation -- a new approach to adapt recurrent neural models on the task of the multi-label classification while avoiding the problem of exposure bias.  
We argue that, in multi-label emotion classification, it is necessary to consider the correlations that exist between the labels, and applying LVC transformation is ideal in this scenario. 
Therefore, we propose a model named Seq2Emo, which not only makes use of many state-of-the-art pre-trained models but also is tailored to adapt to the LVC transformation.  Seq2Emo uses a  bi-directional decoding scheme to capture the relations of both directions. 
Our experiments reveal that the proposed Seq2Emo model performs
better on the datasets containing higher percentages of multi-labeled examples.
It also indicates that our proposed model scales better on the amount of correlations between the labels.


However, we also notice some limitations of the proposed system. The LVC transformation needs $k$ decoding steps for the label set $\mathcal{L}$ of size $k$. If the label sets are very large, for example, RCV1-V2 \cite{lewis2004rcv1} which has 103 labels, the decoding length might be too long for an RNN based model to capture the long distance dependency. In addition, the time complexity of the model is also linearly related to $k$, which potentially makes LVC based models hard to be scaled on the MLC tasks with a large number of distinct labels. 
For emotion mining and other practical MLC problems, the number of distinct labels is typically reasonably small.

\section{Supplementary Material}
\beginsupplement
\subsection{Collection of BMET}

We collect data from scratch mostly following the procedure described in \cite{abdul2017emonet}. The general idea is to find specific hashtags in tweets and assume they are self-annotated \cite{mintz2009distant}. We collect more than 4 billion tweets dated from 2011 to 2018 from Archive.org\footnote{\url{https://archive.org/details/twitterstream}}. From the tweets, we first filter out those that are not English. Then, we define an overall of 46 hashtags to extract 9 emotions (same emotion categories as CBET) to further filter the data keeping only those containing at least two emotions. Based on the emotion distribution, we removed the emotion \emph{love} because it occurs in about  87\% of the tweets. We also removed the emotion \emph{guilt} and 
\emph{disgust} 
as
they together appear in less than 1\% of the tweets. 
Table \ref{tab:bmet_hashtags} enumerates these hashtags. We then remove the hashtags that are used for crawling. To reduce the computational cost, we only use the tweets that have a length ranging between 3 and 50 words. For pre-processing, we used the tool provided by \cite{baziotis-pelekis-doulkeridis:2017:SemEval2}. In order to retain the semantic meanings of the emojis, we first convert the emojis to their textual aliases and then replace the deliminator such as  the ``:'' and ``\_'' with spaces. In order to make the dataset balanced, we first divide the datasets into two portions: multi-labeled only and single-labeled only. We then calculate the label distribution of the multi-labeled part and fill it up by randomly sampling instances from the single-labeled part. 

\begin{table}[ht]
	\caption{Hashtags used to extract the BMET dataset}
	\label{tab:bmet_hashtags}
	\centering		
	\resizebox{0.45\textwidth}{!}{
		\begin{tabular}{l|l} \hline \hline 
			\textbf{Emotion} & \textbf{List of Hashtags}  \\\hline
			anger & \#anger, \#angry, \#rage, \#pain \\
			\hline
			fear & \#fear, \#anxiety, \#horror, \#horrific \\
			\hline
			joy & \#happy, \#joy, \#like, \\ 
			& \#happiness,  \#smile, \#peace, \\
			& \#pleased, \#satisfied, \#satisfying  \\
			\hline
			sadness & \#sad, \#sadness, \#depression, \\
			&  \#depressed, \#alone \\
			\hline
			surprise & \#surprise, \#amazing, \#awesome, \\
			& \#fascinate, \#fascinating, \#incredible, \\
			&  \#marvelous, \#prodigious, \#shocking,   \\
			& \#unbelievable, \#stunning, \#surprising \\
			\hline
			thankfulness & \#thankfulness, \#gratefulness,  \#gratitude, \\
			&   \#kindness, \#thankful, \#thanks, \#gratful \\ \hline \hline
	\end{tabular}}
	
\end{table}

We show several examples with multiple labels from the BMET on Table \ref{tab:eg_bmet}. The hashtags for the six emotions which are used to collect BMET datasets are shown in Table \ref{tab:bmet_hashtags}.

\begin{table*}[ht!]
	\caption{Some examples from BMET dataset. Note that the hashtags expressing emotion labels are removed from the text.}
	\label{tab:eg_bmet}
	\resizebox{0.95\textwidth}{!}{
		\centering	
		\begin{tabular}{c|l} \hline  \hline  
			Emotions & Tweet   \\ \hline 
			surprise, joy &  The moon looks \#amazing :) it's hiding behind a thin curtain of clouds :) \#smile  \\
			anger, sadness & What am I doing wrong??? Just cant seem to make it happen... \#confusion \#anger \#sadness \\
			thankfulness, joy & Even though I tend to get stressed and overreact, I am actully pretty content at how my life is going \#thankful \#happy \\
			thankfulness    &	Nicely done class of 2014! Thank you for helping make @FSU\_HESA   students' experiences better! \#grateful \#LifeNet \textless url\textgreater
			\\   \hline  \hline 
	\end{tabular}}
\end{table*}

\subsection{SL, and SLD on single label emotion classification}
\label{ap:1}
In our experiments, Seq2Emo is outperformed by binary-SLD model on the CBET dataset. In this section, we show that SLD is a strong baseline that is able to achieve great performance on single label emotion classification task.  Compared to binary-SL and binary-SLD, only one single FC layer with \emph{Softmax} regularizer is used to map the output of the encoders to the space of labels. 
In the state-of-the-art emotion classification work by \cite{ZhangFSZWY18}, they use lexicons to estimate the emotion distribution and use multi-task learning to train a CNN text classifier (MTCNN). We run three single labeled emotional datasets against their work: ISEAR \cite{scherer1994evidence}, TEC \cite{mohammad2012emotional} and only single labeled instances in CBET dataset. 
We compare MTCNN with SL, SLD models and three other non- deep learning models (Na\"{i}ve Bayes, Random Forest, and SVM) which use bag-of-word features to represent words. It needs to be mentioned that in the work of \cite{ZhangFSZWY18}, the performance of MTCNN is measured using the averages scores of the evaluation set of the 10-fold cross validation, whereas we run the numbers on a held-out test (10\% of the original dataset). The results are shown in Figure \ref{tab:single_emo_results}, from which we notice that on both TEC and CBET datasets, SLD outperforms MTCNN by a large margin, but MTCNN outperforms SLD on ISEAR datast. ISEAR has only 7,666 instances in total, whereas TEC has 21,051 instances and CBET-Single 76,860 instances. The fact that MTCNN has better scores on the ISEAR maybe because the dataset is too small to be generalized on the held-out test set. 

\begin{table}[ht]
	\caption{Results on single label emotion classification}
	\label{tab:single_emo_results}
	\centering
	\resizebox{0.45\textwidth}{!}{
		\begin{tabular}{l|cccc}  \hline \hline
			\multirow{2}{*}{Models}    & & ISEAR & &  \\  \cline{2-5}
			&   Macro P.  & Macro R. & Macro F1  &  Micro F1  \\  \hline
			Na\"{i}ve Bayes 	 & 	0.5351	 & 	0.536	 & 	0.5338	 & 	0.5359 \\ 
			Random Forest	 & 	0.5460	 & 	0.5467	 & 	0.5363	 & 	0.5463 \\ 
			SVM	 & 	0.5010	 & 	0.5061	 & 	0.5004	 & 	0.5059 \\ 
			Bi-LSTM	 & 	0.6167	 & 	0.6132	 & 	0.6068	 & 	0.6128 \\ 
			MTCNN$^*$    & \textbf{0.6711}    & \textbf{0.6691} &   \textbf{0.6680} &  --    \\
			SL	 & 	0.6420	 & 	0.6261	 & 	0.6296	 & 	0.6258  \\ 
			SLD	 & 0.6514	 & 	0.6523	 & 	0.6451 & 	0.6519  \\   \hline 
			\multirow{2}{*}{Models}  & & TEC & &  \\   \cline{2-5}
			&   Macro P.  & Macro R. & Macro F1  &  Micro F1  \\  \hline
			Na\"{i}ve Bayes 	 & 	0.5477	 & 	0.4841	 & 	0.5008	 & 	0.6035 \\ 
			Random Forest	 & 	0.5831	 & 	0.4024	 & 	0.4333	 & 	0.5741 \\ 
			SVM	 & 	0.5369	 & 	0.4969	 & 	0.5079	 & 	0.5935 \\ 
			Bi-LSTM	 & 	0.5904	 & 	0.5345	 & 	0.5555	 & 	0.6505 \\ 
			MTCNN$^*$   & 0.6210 &  0.5257  & 0.5694 &  --    \\
			SL 	 & 	0.6207	 & 	0.5568	 & 	0.5759	 & 	0.6543  \\ 
			SLD	 & 	\textbf{0.6522}	 & 	\textbf{0.5817}	 & 	\textbf{0.6034}	 & 	\textbf{0.6781}  \\  \hline 
			\multirow{2}{*}{Models} & & CBET-Single & &  \\  \cline{2-5} 
			&   Macro P.  & Macro R. & Macro F1  &  Micro F1  \\  \hline
			Na\"{i}ve Bayes    & 0.5277 & 0.5245   & 0.5218 & 0.5218  \\
			Random Forest  & 0.5188 & 0.5128   & 0.5111 & 0.5218 \\
			SVM            & 0.5288 & 0.5258   & 0.5263 & 0.5258 \\
			Bi-LSTM           & 0.5985 & 0.5993   & 0.5977 & 0.5993 \\
			MTCNN$^*$   & 0.6120 & 0.6158 & 0.6139 &  --    \\
			SL        & 0.6102 & 0.6107   & 0.6092 & 0.6107 \\
			SLD      & \textbf{0.6312} & \textbf{0.6324}   & \textbf{0.6302}  & \textbf{0.6324}  \\ \hline \hline
	\end{tabular}}
\end{table}

\bibliographystyle{aaai}
\bibliography{aaai_clean.bib}

\end{document}